\documentclass[10pt,twocolumn,letterpaper]{article}

\usepackage{cvpr}              
\makeatletter
\@namedef{ver@everyshi.sty}{}
\makeatother

\usepackage{graphicx}
\usepackage{amsmath}
\usepackage{amssymb}
\usepackage{booktabs}

\usepackage[table]{xcolor}
\usepackage{multirow}
\usepackage{tikz}
\usepackage[accsupp]{axessibility} 

\usepackage[pagebackref,breaklinks,colorlinks]{hyperref}

\newcommand{\figoverallarch}[1]{
\begin{figure}[t]
  \centering
   \includegraphics[width=0.9\linewidth]{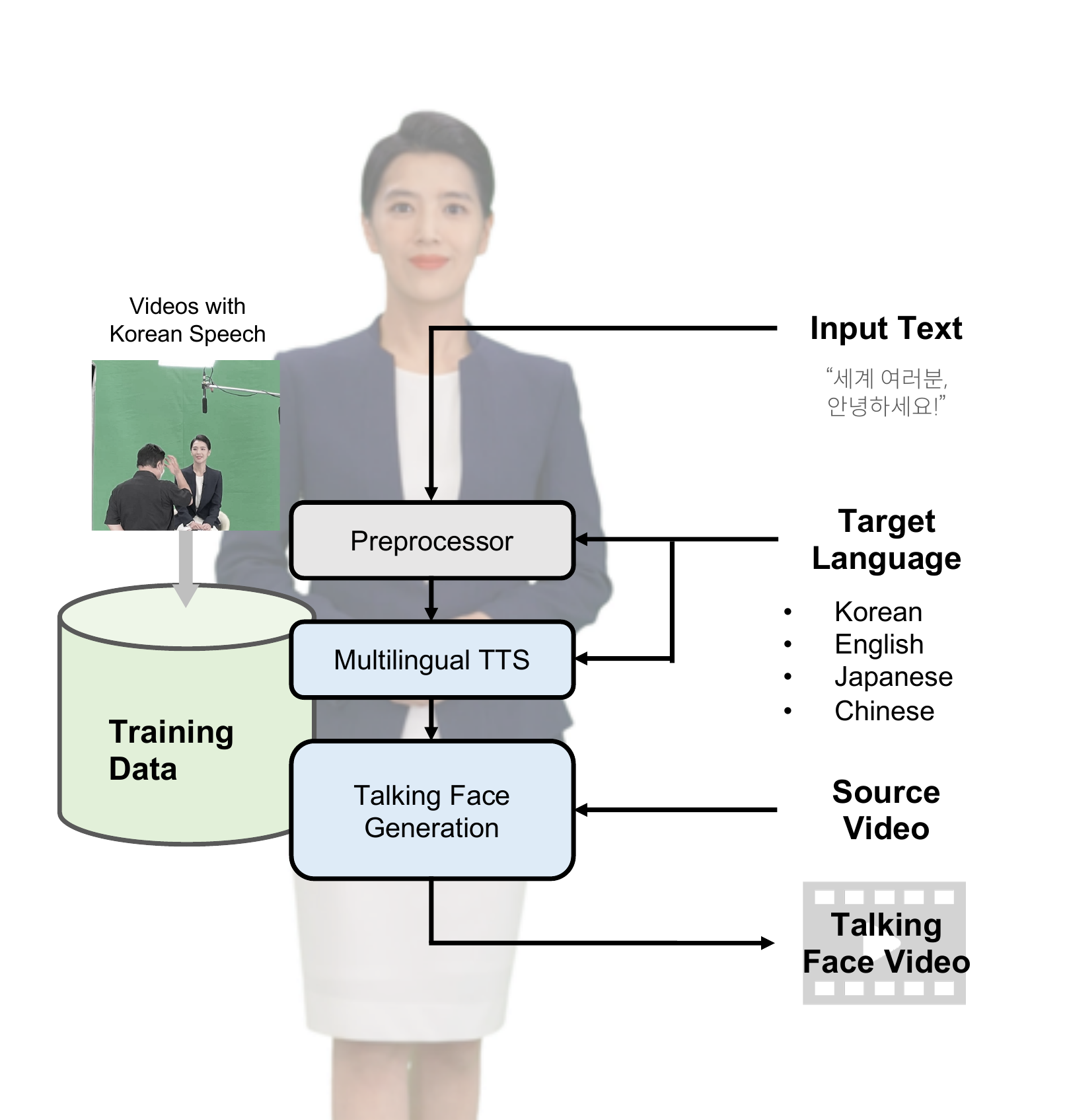}
   \caption{#1}
   \label{fig:overallarch}
\end{figure}
}

\newcommand{\figmodulearch}[1]{
\begin{figure*}[ht]
  \centering
  \includegraphics[width=0.575\linewidth]{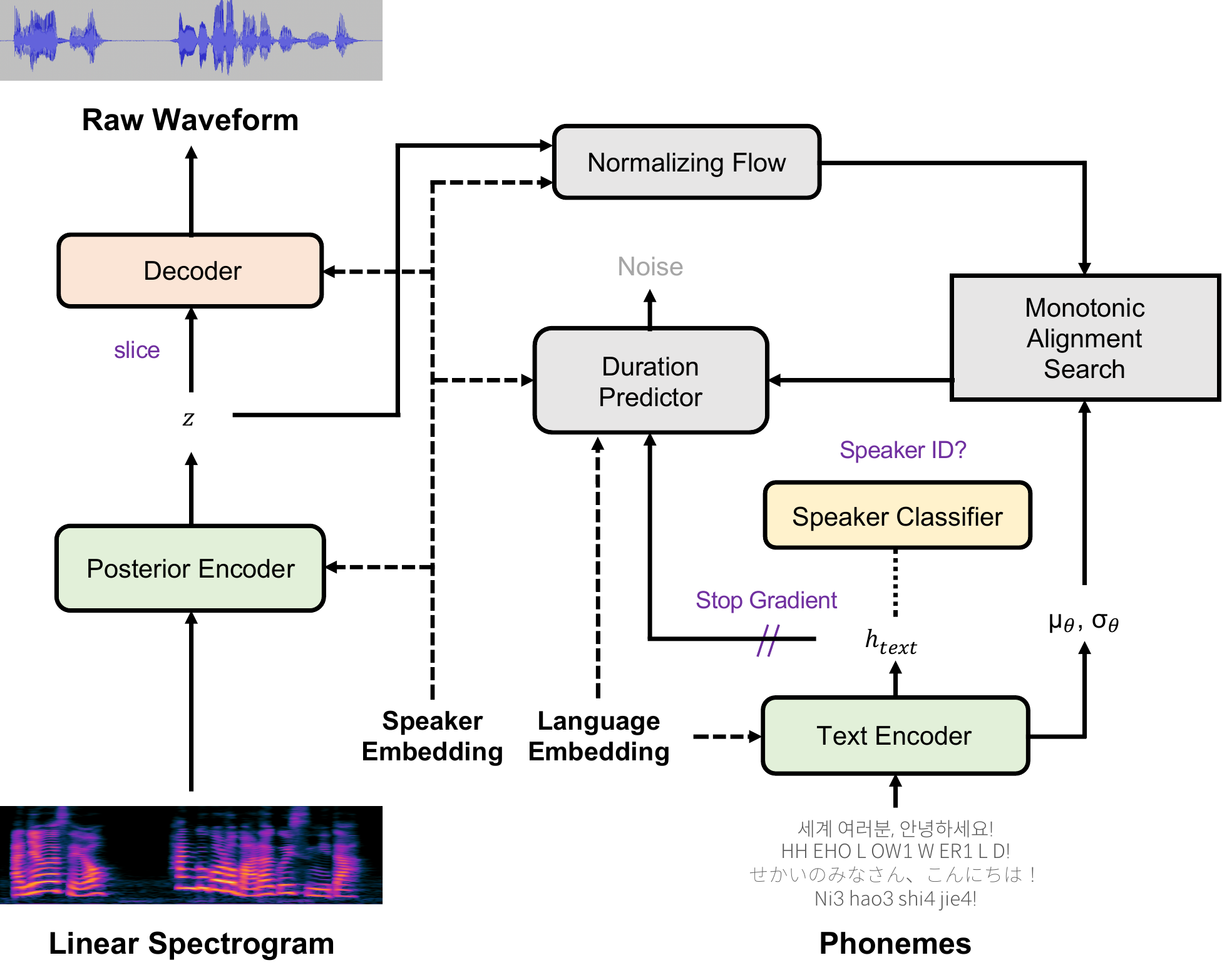}\hfill
  \includegraphics[width=0.375\linewidth]{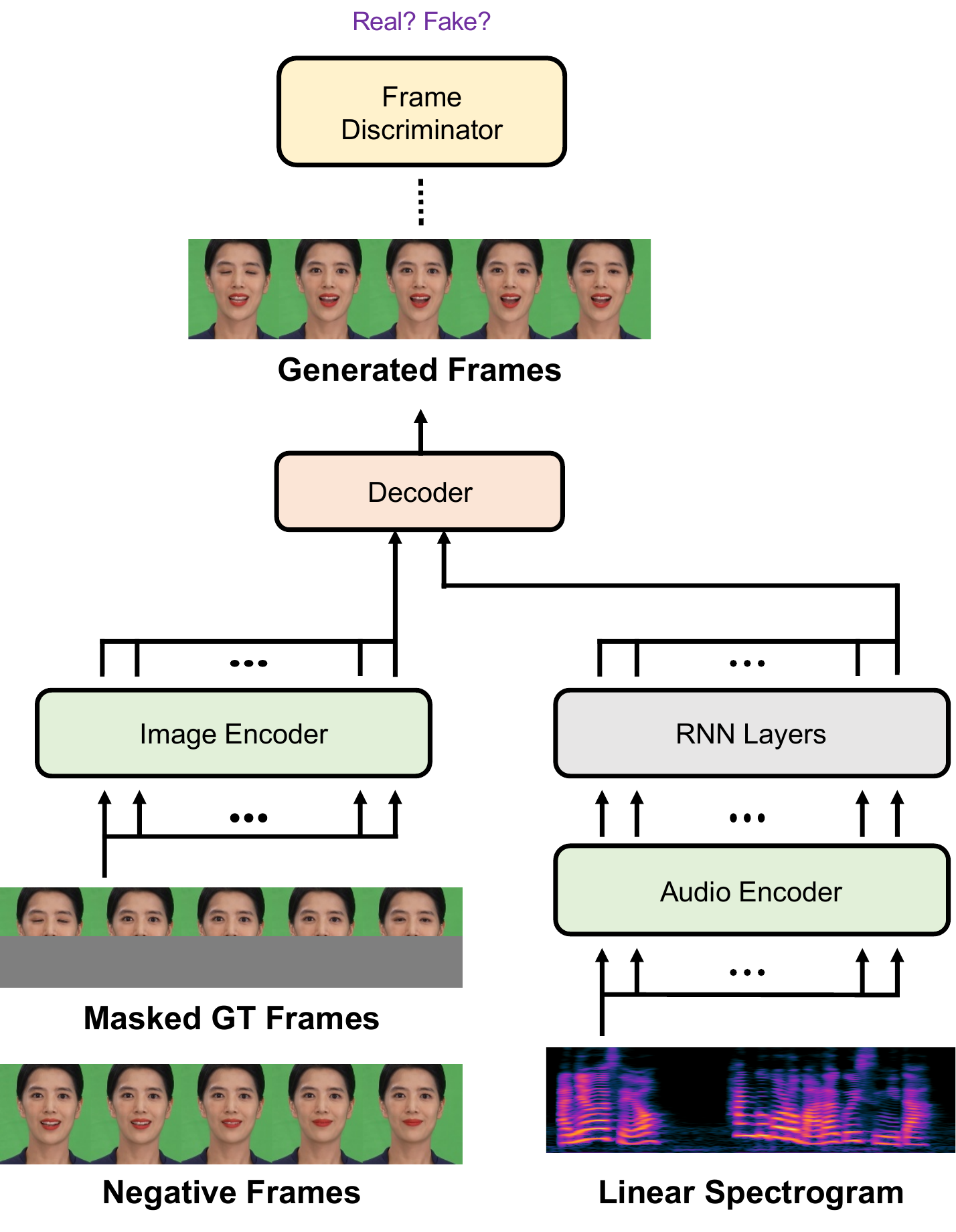}
   \makebox[0.575\linewidth]{\footnotesize {
   (a) Training architecture of multilingual TTS
   }}\hfill
   \makebox[0.375\linewidth]{\footnotesize {
   (b) Training architecture of talking face generation
   }}

   \caption{#1}
   \label{fig:modulearch}
\end{figure*}
}

\newcommand{\figfaceresult}[1]{
\begin{figure*}[ht]
    \centering
    \setlength{\tabcolsep}{3pt}
    \newcommand{\facewidth}{0.19\linewidth}
    \newcommand{\timelabel}{\centering
    \makebox[\facewidth]{\footnotesize0.2 s}%
    \makebox[\facewidth]{\footnotesize0.4 s}%
    \makebox[\facewidth]{\footnotesize0.6 s}%
    \makebox[\facewidth]{\footnotesize0.8 s}%
    \makebox[\facewidth]{\footnotesize1.0 s}
    }
    \newcolumntype{x}{>{\centering\arraybackslash\vspace{0pt}}m{0.28\linewidth}}
    \begin{tabular}{cxxx}
    \toprule
    \multirow{2}{*}{
    \begin{tikzpicture}
        \draw [black,dotted] (0.000\linewidth, 0.05\linewidth) -- (0.10\linewidth, 0.01\linewidth);
        \node [draw=none,anchor=south west,inner sep=0] (A) at (0.000\linewidth, 0.000\linewidth) {\footnotesize {Languages}};
        \node [draw=none,anchor=north west,inner sep=0] (B) at (0.050\linewidth, 0.05\linewidth) {\footnotesize {Models}};
    \end{tikzpicture}} & {(a) \wavtolip (Unseen)} & {(b) \wavtolip (Seen)} & {\textbf{(c) Ours}} 
    \\
    {} & {\timelabel} & {\timelabel} & {\timelabel}
    \\ \midrule
    Korean &
    \includegraphics[width=\facewidth]{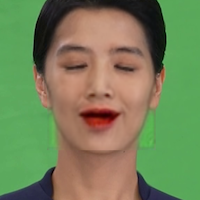}%
    \includegraphics[width=\facewidth]{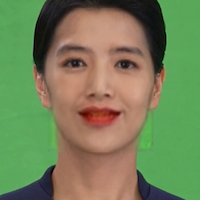}%
    \includegraphics[width=\facewidth]{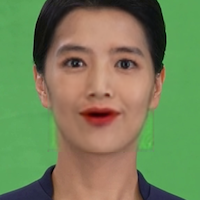}%
    \includegraphics[width=\facewidth]{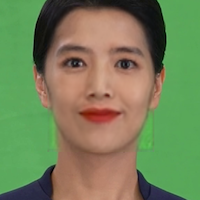}%
    \includegraphics[width=\facewidth]{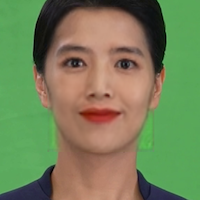}
    &
    \includegraphics[width=\facewidth]{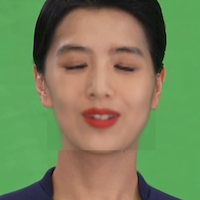}%
    \includegraphics[width=\facewidth]{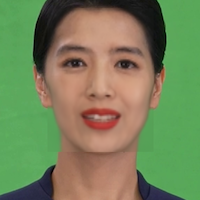}%
    \includegraphics[width=\facewidth]{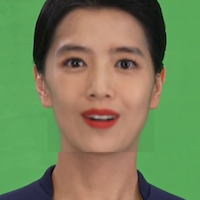}%
    \includegraphics[width=\facewidth]{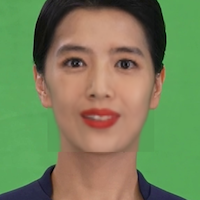}%
    \includegraphics[width=\facewidth]{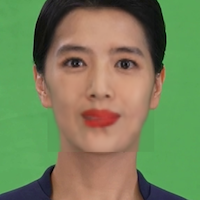}
    &
    \includegraphics[width=\facewidth]{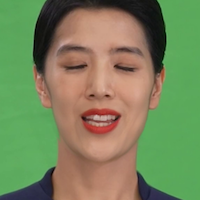}%
    \includegraphics[width=\facewidth]{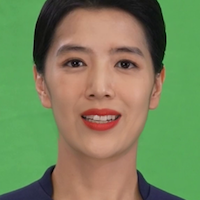}%
    \includegraphics[width=\facewidth]{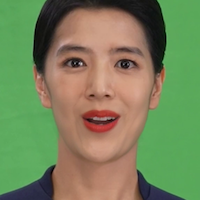}%
    \includegraphics[width=\facewidth]{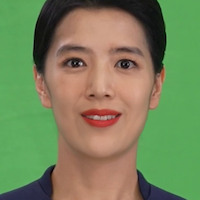}%
    \includegraphics[width=\facewidth]{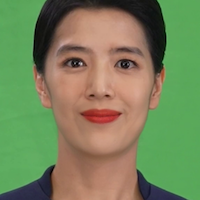}
    \\
    Chinese &
    \includegraphics[width=\facewidth]{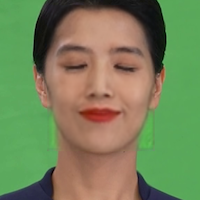}%
    \includegraphics[width=\facewidth]{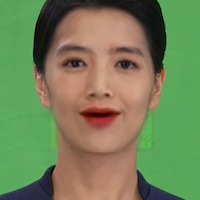}%
    \includegraphics[width=\facewidth]{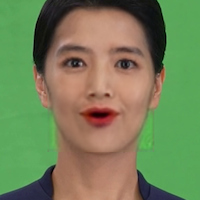}%
    \includegraphics[width=\facewidth]{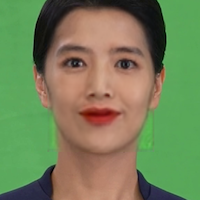}%
    \includegraphics[width=\facewidth]{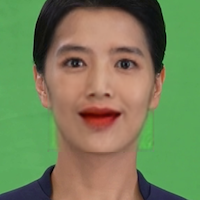}
    &
    \includegraphics[width=\facewidth]{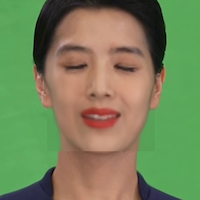}%
    \includegraphics[width=\facewidth]{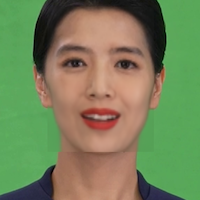}%
    \includegraphics[width=\facewidth]{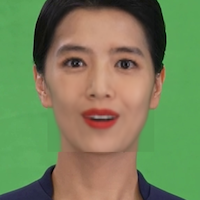}%
    \includegraphics[width=\facewidth]{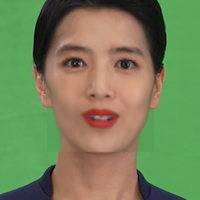}%
    \includegraphics[width=\facewidth]{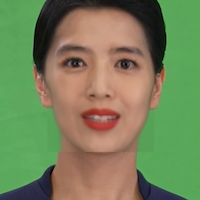}
    &
    \includegraphics[width=\facewidth]{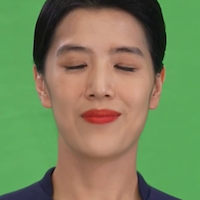}%
    \includegraphics[width=\facewidth]{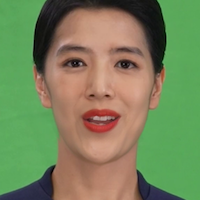}%
    \includegraphics[width=\facewidth]{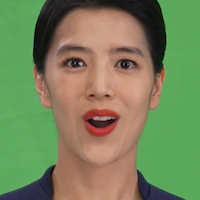}%
    \includegraphics[width=\facewidth]{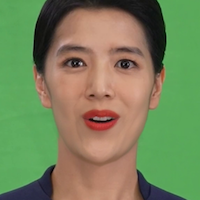}%
    \includegraphics[width=\facewidth]{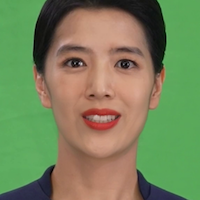}
    \\
    English &
    \includegraphics[width=\facewidth]{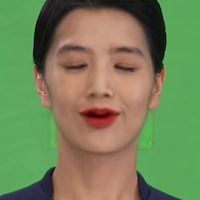}%
    \includegraphics[width=\facewidth]{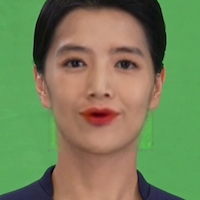}%
    \includegraphics[width=\facewidth]{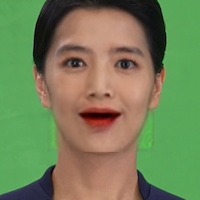}%
    \includegraphics[width=\facewidth]{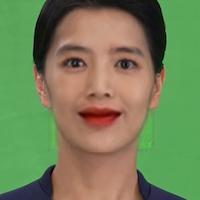}%
    \includegraphics[width=\facewidth]{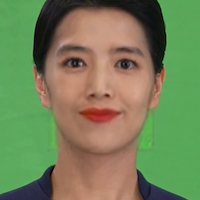}
    &
    \includegraphics[width=\facewidth]{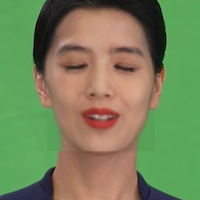}%
    \includegraphics[width=\facewidth]{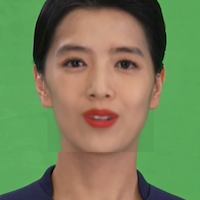}%
    \includegraphics[width=\facewidth]{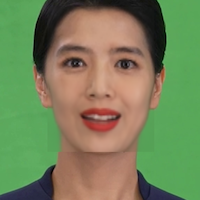}%
    \includegraphics[width=\facewidth]{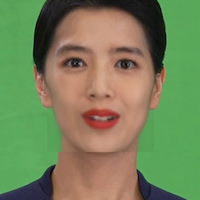}%
    \includegraphics[width=\facewidth]{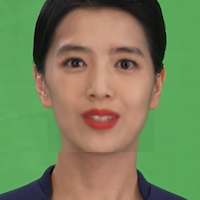}
    &
    \includegraphics[width=\facewidth]{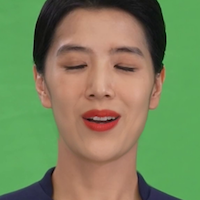}%
    \includegraphics[width=\facewidth]{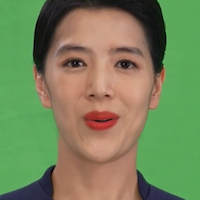}%
    \includegraphics[width=\facewidth]{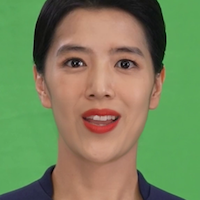}%
    \includegraphics[width=\facewidth]{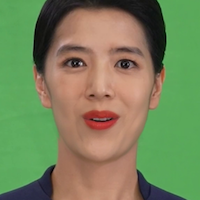}%
    \includegraphics[width=\facewidth]{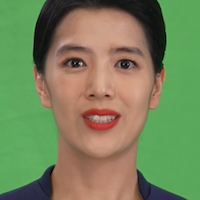}
    \\
    Japanese &
    \includegraphics[width=\facewidth]{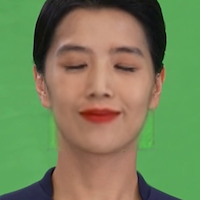}%
    \includegraphics[width=\facewidth]{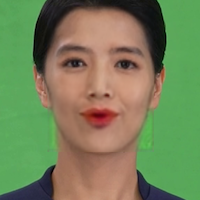}%
    \includegraphics[width=\facewidth]{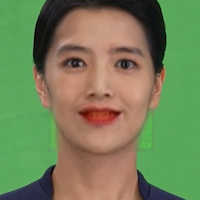}%
    \includegraphics[width=\facewidth]{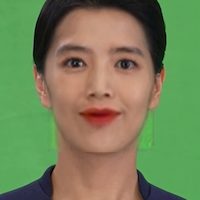}%
    \includegraphics[width=\facewidth]{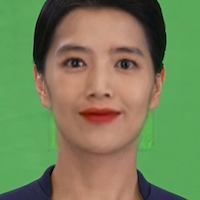}
    &
    \includegraphics[width=\facewidth]{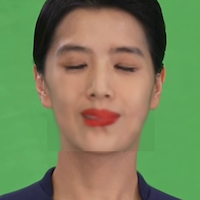}%
    \includegraphics[width=\facewidth]{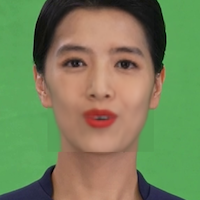}%
    \includegraphics[width=\facewidth]{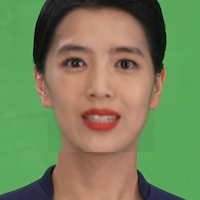}%
    \includegraphics[width=\facewidth]{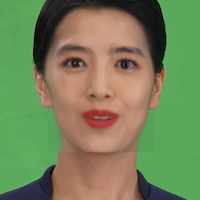}%
    \includegraphics[width=\facewidth]{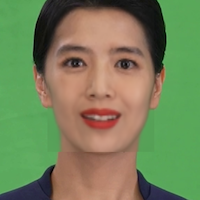}
    &
    \includegraphics[width=\facewidth]{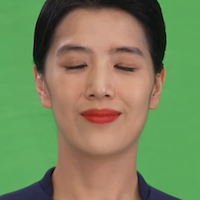}%
    \includegraphics[width=\facewidth]{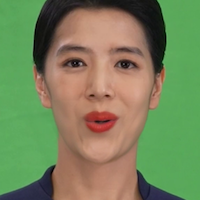}%
    \includegraphics[width=\facewidth]{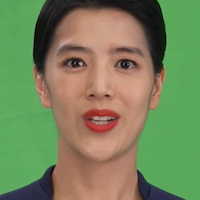}%
    \includegraphics[width=\facewidth]{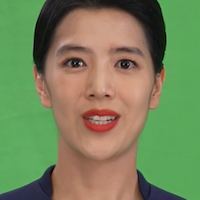}%
    \includegraphics[width=\facewidth]{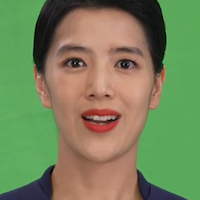}
    \\ \bottomrule
    \end{tabular}
    \caption{#1}
    \label{fig:faceresult}
\end{figure*}
}

\newcommand{\tabspeed}[1]{
\begin{table}[t]
\centering
\renewcommand{\arraystretch}{1.2} 
\newcolumntype{x}{>{\centering\arraybackslash\hspace{0pt}}m{0.15\linewidth}}
\footnotesize{
\begin{tabular}{|>{\centering\arraybackslash\hspace{0pt}}m{0.175\linewidth}|lm{0.175\linewidth}|x|x|}
\hline
\multicolumn{3}{|c|}{Stage} & Time (s) & \fps
\\ \hline
TTS & (a) & Total
    & 0.0655        & -                     %
\\ \hline
\multirow{5}{5em}{\centering{}Talking Face Generation}
    & (b.1) & Face\newline Generation       %
    & 0.3964        & 63.1                  %
\\ \cline{2-5}
    & (b.2) & Video\newline Encoding        %
    & 0.2981        & -                     %
\\ \cline{2-5}
    & (b) & Total                           %
    & 0.6945        & 36.0                  %
\\ \cline{2-5}
& \multicolumn{2}{c|}{(a) + (b.1)}
    &\textbf{0.4619}& \textbf{54.1}         %
\\ \cline{2-5}
& \multicolumn{2}{c|}{(a) + (b)}
    &\textbf{0.7600}& \textbf{32.9}         %
\\ \hline
Overall System & (c) & Total
    &\textbf{0.8251}& \textbf{30.3}         %
\\ \hline
\end{tabular}\vspace{1.5mm}
}
\caption{#1}
\label{tab:speed}
\end{table}
}
\newcommand*\samethanks[1][\value{footnote}]{\footnotemark[#1]}

\newcommand{\fps}{fps\xspace}
\newcommand{\wavtolip}{Wav2Lip\xspace}

\usepackage[capitalize]{cleveref}
\crefname{section}{Sec.}{Secs.}
\Crefname{section}{Section}{Sections}
\Crefname{table}{Table}{Tables}
\crefname{table}{Tab.}{Tabs.}

\def\cvprPaperID{8} 
\def\confName{CVPR Demo Track (Round1)}
\def\confYear{2022}

\begin{document}

\title{Talking Face Generation with Multilingual TTS}

\author{
Hyoung-Kyu Song\thanks{~indicates equal contribution.}$~~^1$$^2$,
Sang Hoon Woo\samethanks$~~^1$,
Junhyeok Lee$^1$,
Seungmin Yang$^1$$^2$,\\
Hyunjae Cho$^1$$^3$,
Youseong Lee$^1$$^3$,
Dongho Choi$^3$,
Kang-wook Kim$^3$
\\
$^1$MINDsLab Inc., South Korea\\
$^2$KAIST, South Korea\hspace{2mm}
$^3$Seoul National University, South Korea\\
{\tt\small \{hksong35, shwoo, jun3518, myaowng2, chohyunjae, staru\}@mindslab.ai}\\
{\tt\small \{dongho.choi, full324\}@snu.ac.kr}}

\maketitle

\begin{abstract}
   In this work, we propose a joint system combining a talking face generation system with a text-to-speech system that can generate multilingual talking face videos from only the text input.
Our system can synthesize natural multilingual speeches while maintaining the vocal identity of the speaker, as well as lip movements synchronized to the synthesized speech.
We demonstrate the generalization capabilities of our system by selecting four languages (Korean, English, Japanese, and Chinese) each from a different language family.
We also compare the outputs of our talking face generation model to outputs of a prior work that claims multilingual support.
For our demo, we add a translation API to the preprocessing stage and present it in the form of a neural dubber so that users can utilize the multilingual property of our system more easily.
\end{abstract}


\section{Introduction}
\label{sec:intro}

Talking face generation, a task of synthesizing a face video where the lip is synchronized with the input speech, is one of the most popular research topics in neural video generation. 
When combined with a text-to-speech (TTS) system, the joint system allows users to create a talking video with only a text input and has potential applications in news broadcasting, virtual lectures, and digital concierge. 
Expanding the task to support multiple languages would significantly reduce the amount of effort required to widen the target audience to the global population. 

\figoverallarch{
An overview of the demonstration. By inputting the text, the language, and the source video, users can make a multilingual talking face video. For the training data, we record two hours of footage of the Korean speaker. 
}

Recent works in talking face generation claim that their models support input speeches in any language \cite{prajwal2020lip,lahiri2021lipsync3d}.
However, we observe that such models fail to generalize to certain input speech languages, \eg, Korean. 
We hypothesize that the robustness of these models depends on the degree of similarity between the training speech language and the input speech language. 
Thus, we will validate the generalization capabilities of multilingual face generation models using speeches of languages from different language families.

For practical applications of multilingual talking face generation, the speaker’s vocal identity should be preserved across different languages. 
Since multilingual speech datasets for desired speakers are often unavailable, the multilingual talking face generation system requires a multilingual TTS model capable of cross-lingual speech synthesis.
While a number of prior works in multilingual TTS discuss cross-lingual speech synthesis, the selection of languages has been underexplored. 
Thus, existing works' ability to perform cross-lingual synthesis across languages from different language families is questionable.

In this work, we propose a multilingual talking face generation system shown in Fig. \ref{fig:overallarch}. We also describe the two models used in the multilingual TTS module and the talking face generation module: a multilingual adaptation of VITS \cite{kim2021conditional} capable of performing cross-lingual speech synthesis while preserving the speaker's vocal identity, and a talking face generation model capable of generating face videos from synthesized speeches, regardless of the language. 

Our contributions in this work are the following: 

\begin{itemize}

    \item 
    We introduce a system that can synthesize talking face videos in four languages (Korean, English, Japanese, and Chinese) for a monolingual speaker.
    \item  
    We build a talking face generation model that is robust to different input speech languages.
    \item 
    Our demonstration can generate \(512 \times 512\) facial image sequences faster than 25 \fps.
\end{itemize}

\section{Related Work}
\label{sec:relatedwork}

\subsection{Text-To-Speech (TTS)}
\label{subsec:relatedwork_tts}

Traditionally, text-to-speech systems have employed a two-stage pipeline, with the models for each stage developed independently from each other. 
The first stage uses an acoustic model \cite{shen2018natural, ren2021fastspeech} to generate an intermediate speech representation, namely mel-spectrogram, based on the input text. 
In the second stage, a vocoder model \cite{prenger2019waveglow, kong2020hifi} converts the speech representation to a raw waveform. 
There have been a number of attempts \cite{ren2021fastspeech,chen2021wavgrad2} to reduce text-to-speech systems to a fully end-to-end process, but they required complex input conditions \cite{ren2021fastspeech} or were slow \cite{chen2021wavgrad2}. 
Recently, Kim \etal \cite{kim2021conditional} proposed VITS, a non-autoregressive end-to-end architecture that surpasses state-of-the-art two-stage models. 

Among speech synthesis related studies, some works have focused on multilingual text-to-speech models with cross-lingual capabilities.  
Zhang \etal \cite{zhang19learning} first proposed the use of domain adversarial training in multilingual text-to-speech training to mitigate the speaker dependency in text representations. However, the quality of the generated speech was highly dependent on the source speaker’s language and the target speech language. 
Maiti \etal \cite{maiti2020generating} utilized bilingual speaker data to compute the modification between different languages in the speaker embedding space. 
No prior works explored cross-lingual speech synthesis with languages from different language families.

\subsection{Talking Face Generation}
\label{subsec:relatedwork_facegen}

Recent works in talking face generation have focused on generalizing their models to any vocal and visual identity \ie from any input speech to any target face. 
Among such works, some have successfully built models that generate videos with realistic faces, using a GAN-based approach  \cite{vougioukas2018end, prajwal2020lip}. 
Vougioukas \etal \cite{vougioukas2018end} proposed temporal GANs that include recurrent layers in the architecture for temporal consistency. 
Prajwal \etal \cite{prajwal2020lip} suggested the use of modified SyncNet \cite{chung2016out} to determine whether the generated images and the corresponding audio are in sync. 
Both networks described above have low resolution outputs; Vougioukas \etal \cite{vougioukas2018end} outputs $96 \times 96$ images and Prajwal \etal \cite{prajwal2020lip} outputs $96 \times 128$ images. 
The low resolution of the output images also restricts the maximum resolution of the final output video.

Prior talking face generation studies claim that their systems are training language agnostic, \ie, their models can generate videos with speech from any language, regardless of the language of the data the models were trained on. 
While such claims do hold for some languages, the quality of the outputs tends to degrade significantly when tested on a language from a different family tree than the training language. 
This phenomenon can be observed in the official interactive demo of \cite{prajwal2020lip}.
Such cases often require re-training of the model with data in the desired language.

\figmodulearch{
Training pipeline of the multilingual TTS model and the talking face generation model.
In (a), the multilingual TTS model synthesizes a raw speech waveform from the input text, speaker embedding, and the language embedding.
In (b), the talking face generation model generates a sequence of face images from the source face images and the input speech.
}

\section{Face Generation with Multilingual TTS}
\label{sec:method}

\subsection{Overall Architecture}
\label{subsec:method_overall}

The proposed system consists of three main modules: a preprocessor module, a multilingual TTS module, and a talking face generation module. 
First, the preprocessor module transforms the input text to a phoneme sequence. 
The multilingual TTS module, in turn, generates a raw speech waveform based on the phoneme sequence with the designated speaker identity. 
Subsequently, the talking face generation module synthesizes the final output video with the lip movement synchronized to the input speech. 

\subsection{Preprocessor}
\label{subsec:method_preprocessor}

While the details of the preprocessing step differ for each language, the common objective is to convert raw text into a sequence of phonemes. 
The preprocessor first cleans the text by removing any character or symbol that does not belong to the specified language. 
Then the text goes through a normalization procedure, converting non-verbal texts, \eg, digits, dates, and short forms of words, to their verbalized forms. 
Finally, the preprocessor maps all graphemes in the text to phonemes. 
We use different phoneme sets for each language in our setup; the phoneme sets for Korean, English, Japanese, and Chinese are Hangul, ARPAbet, Hiragana, and Pinyin, respectively. 
For Korean and English, we utilize our in-house grapheme-to-phoneme algorithms.
For Japanese and Chinese grapheme-to-phoneme conversion, we use open source libraries SudachiPy\footnote{\url{https://github.com/WorksApplications/SudachiPy}} and pypinyin\footnote{\url{https://github.com/mozillazg/python-pinyin}}. 
The preprocessor may employ an optional language translation system for downstream applications like a neural dubber. 

\subsection{Multilingual TTS}
\label{subsec:method_multilingualtts}

We use multi-speaker VITS \cite{kim2021conditional} as the base model for the multilingual TTS module. 
To enable multilingual speech synthesis for VITS, we add embeddings for each language and input them to submodules.
Our preliminary experiments showed that injecting the language embedding to the text encoder and the duration predictor yields the best result. 
We will refer to this setting as the multilingual VITS. 

\subsection{Talking Face Generation}
\label{subsec:method_facegen}

For talking face generation, we use our internal talking face generation model with output image resolution of $512 \times 512$. 
Our face generation model comprises three components: an image encoder, an audio encoder, and a decoder. 
The image encoder and the audio encoder extract features from the input image and the audio, respectively. 
We concatenate the extracted features from both encoders and feed them to the decoder, which then generates lip-synchronized face images.
All three components of our module are primarily CNN-based models. 
The audio encoder that employs additional RNN layers to maintain temporal consistency. 
Thus, our module can generate image sequences with minimal autoregressive computation, reducing the network latency.

\subsection{Dataset Collection}
\label{subsec:method_dataset}

\figfaceresult{Comparison of the output image sequences. For (a), we use the official interactive demo of \wavtolip. For (b), we train a model with the same target identity dataset as (c). All output images are cropped using the same method as our system's facial region cropping. The image sequences for each model-language pair were taken every 0.2 seconds over 1.0 second. Note that all models used the same input speech and the source video.}

For the training data of our target identity, we record two hours of footage of the target speaker speaking in Korean. 
The recorded video and audio data are resampled to 25 \fps and 22050 Hz, respectively.

We only use the facial regions from the image frames as the input to the talking face generation model. 
To extract the facial region from a image region, we first estimate facial landmarks with a pre-trained model provided by \cite{guo2020towards}. 
A prior study \cite{vougioukas2018end} cropped the facial regions tightly around the detected faces. 
However, we find that if the entire head is not included in the cropped region, the final output image contains borderline discontinuity. 
To mitigate this issue, we expand the facial regions to include the entire head with extra margins.

For the multilingual VITS training data, we use a mix of our internal datasets and open source datasets, in addition to the target speech data.
Our internal datasets comprise 28 hours of Korean speech from 25 speakers and 13 hours of English speech from 13 speakers. We also include several open source text-to-speech datasets as a part of our train set: the Korean Single Speaker Speech dataset \cite{kss} for Korean, the LJ Speech dataset \cite{ljspeech} and the LibriTTS dataset \cite{zen2019libritts} for English, the Voice Actress corpus\footnote{\url{http://voice-statistics.github.io/}}, the JSUT dataset \cite{sonobe2017jsut}, and the JVS dataset \cite{takamichi2019jvs} for Japanese, and the AISHELL-3 dataset \cite{shi2021aishell} for Chinese. Overall, our train set includes speech from 472 speakers totaling 206 hours.

\subsection{Training Details}
\label{subsec:method_training}

The training procedure for the multilingual VITS remains largely the same as the original VITS training \cite{kim2021conditional}. We employ a couple of techniques on top of the original VITS training to improve the quality of multilingual speech synthesis. 
First, we apply domain adversarial training \cite{ganin2016domain} to minimize the speaker information leakage to the encoded text representations. 
Following Zhang \etal \cite{zhang19learning}, we add a speaker classifier with a gradient reversal layer after the text encoder. 
For the classification loss scale factor $\lambda$, we follow the schedule from \cite{ganin2016domain}. 

During the initial experiments, we observed that feeding the speaker embedding directly to the duration predictor degrades the quality of the output speech. 
We hypothesize that the unseen combination of language embedding and speaker embedding introduces instability. 

To resolve this issue, we add a regularization term to the loss, such that the mean of all speaker embeddings are pushed towards the zero vector. 
During inference, if the speaker’s original language does not match the input language, we use the zero vector in place of the speaker embedding, similarly to the use of the zero vector as the residual encoding in Zhang \etal \cite{zhang19learning}.

In training our talking face generation model, we follow Prajwal \etal \cite{prajwal2020lip} and use masked ground truth frames and negative frames, \ie, frames from different part of the video.
In addition, we apply various augmentations, \eg, translation, rotation, zoom in/out to the facial region, instead of normalizing the facial regions to minimize the spatial variance. 
This drives the generator to be more robust to the rotation of the neck or the size ratio of the face to the image when generating the face.
We also adopt adversarial training with multi-scale discriminator \cite{wang2018pix2pixHD} as the auxiliary training to improve the perceptual visual quality.

\section{Experiments}
\label{sec:experiments}

In Fig. \ref{fig:faceresult}, we compare the outputs from our model to the outputs from two versions of the \wavtolip model. 
For both versions of \wavtolip, we observe a number of artifacts in the output images that degrade the overall image quality. 
First, the borderlines of the bounding box are clearly visible near the chin in the output images. 
Also, the generated images from both models are fuzzy and lack details compared to the outputs from our model. 
In \wavtolip (Unseen) outputs, we observe that the inside of the mouth is filled with black in most frames, erasing out details including the teeth.
\wavtolip (Seen) outputs display different artifacts; when the lips are closed, the border between the upper and the lower lips disappears.
We believe that the use of adversarial training forces our model to generate more fine-grained details, resulting in higher quality image overall.

We also report the characteristics of the model outputs for out-of-training-language speeches. 
The outputs from \wavtolip (Unseen) show abrupt lip movements for non-English speeches, characterized by sudden closing of the mouth. 
On the other hand, the outputs from \wavtolip (Seen) exhibit a very different behavior for out-of-training-language speeches; in \wavtolip (Seen) output images, the lip movements are minimal compared to the outputs from the other two models. 
Our model, on the contrary, produces a more gradual and diverse lip movements regardless of the input speech language. 
This shows that our training method builds a model more robust to different input speeches.

\tabspeed{
Throughputs for each stage of the system.
Time values in the second column are measured for generating one second audio or video.
Note that the original video is rendered with 25 \fps.
}

After generating the facial image sequences, the images are merged into the source FHD video before the end users receive them. 
Specifically, we replace the facial regions of the source FHD video frames with the generated face image sequences. 
We measure the end-to-end latency of the system as the duration between the user request and the service response containing the final video. 
We deploy our system on a desktop with AMD Ryzen 7 3800XT and Nvidia GeForce RTX 3080 and measure the speed of our system. 
As shown in Table \ref{tab:speed}, the speed of the entire system is faster than real-time. 
Note that the video encoding time is measured for reference videos that are loaded into the system beforehand; if the system were to support custom reference videos, \ie, user-inputted reference videos, the system will require additional preprocessing time.

In this demo, we choose MP4, a universal file format for videos, as the output format so that the output videos can be played on any device. 
However, the MP4 format does not allow streaming until the entire video is generated, leading to longer latency for the end user. 
Switching to a more streaming-friendly format, such as MKV, would significantly reduce the end-to-end latency and improve its applicability. 

\section{Broader Impact}
\label{sec:impact}

Unlike existing works on talking face generation, the objective of our system is not to support inference on unseen identities. 
Instead, we focus on generating high resolution talking face videos where the target identity is seen during training. 
The proposed system can facilitate the production of video-based media such as virtual newscast or online tutoring. 
Combined with a language translation system, our system allows users to generate four versions of a video, each in a different language, significantly boosting the accessibility of the content.

While the proposed technology does come with benefits, we acknowledge that it can also be used with malicious intents. 
Since the system can generate video based on any text, an adversarial user may attempt to create deepfakes with harmful contents. 
Regarding such vulnerabilities, we first note that the data required to train the proposed system would likely be unobtainable without an agreement with the target identity, substantially reducing the entities that can train the system.  
The system may also employ a content filter, \eg, a hate speech filter to further reduce the risk of generating malicious contents. 
Also, we strongly believe that with proper accountability management involving tracking the use of every generated video, the system can be safely used with minimal liability.

\section{Conclusion}
\label{sec:conclusion}

In this work, we present a robust talking face generation system compatible with multilingual speech from a speech synthesis model. 
We describe a talking face generation model robust to input speech language, as well as techniques to equip a state-of-the-art TTS model with multilingual synthesis capabilities. 
By combining the face generation model and the TTS model, we build a system that can generate talking face videos in four languages without a multilingual parallel dataset. 
We demonstrate our system’s ability to generalize across languages by evaluating with languages from different language families.
We also show that our system is feasible in industrial settings by deploying our demo on a desktop with no external computational resources.
We hope that our system can help content creators improve the accessibility of their works past language barriers. 

{\small
\bibliographystyle{ieee_fullname}
\bibliography{cvpr_final.bbl}
}

\end{document}